\documentclass[a4paper,fleqn]{cas-dc}

\usepackage[numbers]{natbib}
\usepackage{caption}
\newcommand{\pheadWithSpace}[1] {\vspace{2mm}\noindent\textbf{#1.}}

\usepackage{graphics}
\usepackage{amsmath}
\usepackage{amssymb}
\usepackage{epsfig}
\usepackage{subfigure}
\usepackage{array}
\usepackage{multirow}

\usepackage[linesnumbered,ruled,lined]{algorithm2e}
\usepackage{algpseudocode}
\usepackage{algorithmicx}
\usepackage{enumitem}
\usepackage{hyperref} 

\begin{document}
\let\WriteBookmarks\relax
\def\floatpagepagefraction{1}
\def\textpagefraction{.001}
\shorttitle{Joint Semantics and Data-Driven Path Representation for Knowledge Graph Inference}
\shortauthors{Guanglin Niu et~al.}

\title [mode = title]{Joint Semantics and Data-Driven Path Representation for Knowledge Graph Inference}

\author[1]{Guanglin Niu}[style=chinese, orcid=0000-0001-7260-7352]
\author[1,2]{Bo Li}[style=chinese]
\author[1,2,3]{Yongfei Zhang}[style=chinese]\cormark[1]
\author[4]{Yongpan Sheng}[style=chinese]
\author[5]{Chuan Shi}[style=chinese]
\author[2]{Jingyang Li}[style=chinese]
\author[6]{Shiliang Pu}[style=chinese]

\address[1]{Beijing Key Laboratory of Digital Media, School of Computer Science and Engineering, Beihang University, Beijing, China}
\address[2]{State Key Laboratory of Virtual Reality Technology and Systems, Beihang University, China}
\address[3]{Pengcheng Laboratory, Shenzhen, China}
\address[4]{School of Big Data and Software Engineering, Chongqing University, Chongqing, China}
\address[5]{Beijing University of Posts and Telecommunications, Beijing, China}
\address[6]{Hikvision Research Institute, Hangzhou, China}

\cortext[cor1]{Corresponding author\newline\hspace*{0.6cm}\textit{Email address}: \url{yfzhang@buaa.edu.cn} (F. Zhang)}

\begin{abstract}
Inference on a large-scale knowledge graph (KG) is of great importance for KG applications like question answering. The path-based reasoning models can leverage much information over paths other than pure triples in the KG, which face several challenges: all the existing path-based methods are data-driven, lacking explainability for path representation. Besides, some methods either consider only relational paths or ignore the heterogeneity between entities and relations both contained in paths, which cannot capture the rich semantics of paths well. To address the above challenges, in this work, we propose a novel joint semantics and data-driven path representation that balances explainability and generalization in the framework of KG embedding. More specifically, we inject horn rules to obtain the condensed paths by the transparent and explainable path composition procedure. The entity converter is designed to transform the entities along paths into the representations in the semantic level similar to relations for reducing the heterogeneity between entities and relations, in which the KGs both with and without type information are considered. Our proposed model is evaluated on two classes of tasks: link prediction and path query answering task. The experimental results
show that it has a significant performance gain over several different state-of-the-art baselines.
\end{abstract}







\maketitle

\section{Introduction}
\label{intro}


Knowledge graph (KG) incorporates rich multi-relational data in a directed graph structure. Many KGs in the real-world, such as Freebase~\cite{BGF:Freebase}, YAGO~\cite{suchanek2007yago}, WordNet~\cite{Miller:WordNet} and NELL~\cite{Mitchell:nell}, often consist of millions or billions of facts\footnote{\url{https://www.wikidata.org/wiki/Wikidata:Statistics}}, have been established and conducted in a wide variety of real applications such as information extraction\cite{hoffmann2011knowledge,daiber2013improving}, semantic search~\cite{berant2013semantic, berant2014semantic}, question answering\cite{zhang2016question, diefenbach2018wdaqua}, and dialog system~\cite{he2017learning}, to name a few. A KG is essentially a heterogeneous graph in which nodes correspond to \emph{entities} and edges correspond to \emph{relations}. Each directed edge, along with its head entity and tail entity, constitute a triple, i.e., (\emph{head entity}, \emph{relation}, \emph{tail entity}), which is also named as a fact.

\begin{figure}
  \centering
  \includegraphics[scale=0.62]{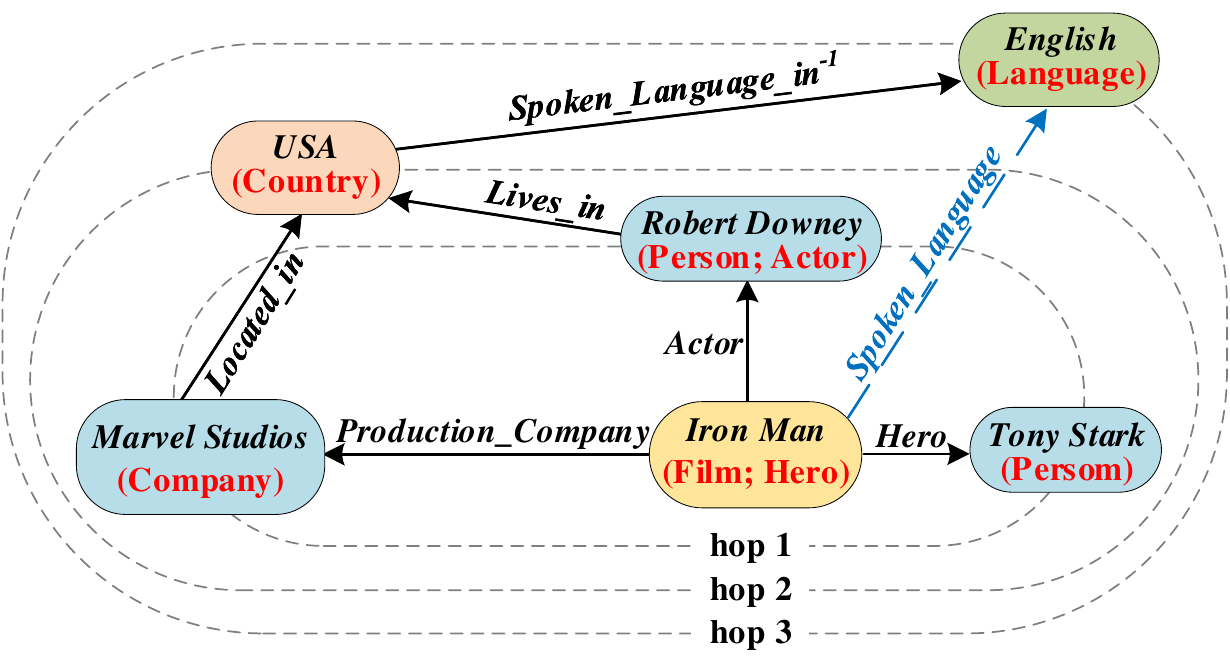}
  \caption{An example of inference on multi-hop paths in the KG. Nodes as entities with the same steps reached from the originator node are in the same color. Entity types are attached to each entity and shown in red color. The superscript ``-1'' of relation means the inverse relation. For this example, the blue dashed directed edge denotes the predicted relation \emph{Spoken\_Language} between \emph{Iron Man} and \emph{A} that is inferred by two 3-hop paths linking \emph{Iron Man} and \emph{English}.}
  \label{fig:ex_multihop}
\end{figure}

KG embedding aims to embed KG elements (i.e., entities and relations) into the latent, low-dimensional, and real-valued vector representations. Over the past decade, it has been attracting substantial attention in academia and industry, and proven to be a powerful technique for KG inference~\cite{KGE:Survey,Analogical, Chami:Hyperbolic}. Many KG embedding techniques have been proposed for the graph structure of KG, including TransE~\cite{Bordes:TransE}, TransH~\cite{Wang:TransH}, TransR~\cite{Lin-a:TransR}, ComplEx~\cite{Trouillon:ComplEx} and ConvE~\cite{Dettmers:CNN}. However, these methods simply consider the direct relations between entities in the KG. It may encounter the issues in some more complicated tasks such as path queries because they often require multiple hops to infer the correct answer, as illustrated by an example in Fig.\ref{fig:ex_multihop}.

An increasing number of researchers have recently paid attention to paths in the KG; its core objective lies in incorporating path information into the KG embedding framework. A path in the KG is typically defined as a sequence consisting of intermediate entities and relations in the knowledge graph to provide more connections between entities. Several typical path-based embedding methods have been proposed, such as PTransE~\cite{Lin-b:PTransE}, RTransE~\cite{RTransE}, DPTransE~\cite{DPTransE}, and RPJE \cite{RPJE}. PTransE develops a path-constraint resource allocation (PCRA) algorithm to measure employs a margin-based loss function, which introduces the relational path representations. RPJE leverages the horn rules to compose paths and create the associations among relations for improving the accuracy and explainability of representation learning. However, these models only consider relations along paths but neglect the semantics implied in entities. For two instances about the paths: \\P1. \small$Donald\ Trump \xrightarrow{PresidentOf} Government \xrightarrow{LocatedIn} USA$ \normalsize and P2. \small$Bill\ Gates$ $\xrightarrow{PresidentOf} Microsoft \xrightarrow{LocatedIn} USA$\normalsize. Although both the paths exactly contain the same relations, these two instances can be inferred respectively as \small(\emph{Donald Trump}, \emph{PresidentOf}, \emph{USA}) \normalsize and \small (\emph{Bill Gates}, \emph{Nationality}, \emph{USA}) \normalsize due to the disparate immediate entities \emph{Government} and \emph{Microsoft}. To jointly consider the relations and entities along paths, Wang~\emph{et al.}~\cite{Wang:PathLSTM} composes entity and relation embeddings in each step for LSTM. Still, entities are only independent individuals, and hard to provide semantics to represent the path. Das~\emph{et al.}~\cite{Das:Chain} replaced each entity over a path by all the types belonging to this entity, but some types have nothing to do with the semantics of the path, such as the type \emph{Hero} of \emph{Iron Man} in Fig.\ref{fig:ex_multihop} cannot contribute to representing a path connecting \emph{Iron Man} and \emph{English}. The latest research in \cite{TypePR} proposes an attentive path ranking approach which selects entity types attentively in representing path patterns. However, such an entity type selection strategy is purely data-driven and this path ranking framework cannot predict entities. Besides, both~\cite{Das:Chain} and~\cite{TypePR} cannot work on the KG without types.

Another research line of path-based reasoning methods is to focus on reinforcement learning (RL). DeepPath~\cite{DeepPath} first learns multi-hop relational paths via RL framework, and then find reliable predictive paths between entity pairs. Das~\emph{et al.}~\cite{Das:MINERVA} and Lin~\emph{et al.}~\cite{Lin:reward-shaping} designed a end-to-end multi-hop KG reasoning model based on the RL. Nevertheless, these RL-based reasoning models spend much more time searching for reliable paths while performing less than the embedding-based methods.

Although the above KG inference methods have achieved preliminary performance improvements on various tasks, such as link prediction and path query answering task, they still suffer from the following challenges:

(1) The model represents the paths in a data-driven pattern manner, which achieved the good generalization but lacked the explainability and sufficient accuracy of the path representation.

(2) Each path is expected to be represented as a relation embedding. Still, the heterogeneity between entities and relations in the path limits the entities to be directly applied to represent the path. Besides, the strategy of employing entity types instead of entities to represent paths cannot be applied to the KG without types.

Based on the above considerations, in this paper, we propose a novel KG embedding paradigm for inference, which achieves jointly semantics and data-driven path representations. Unlike previous path-based approaches in a data-driven fashion, our model attempts to introduce a joint semantics and data-driven mechanism to balance explainability and generalization of the path representation process. The horn rules are exploited to compose the original paths consisting of entities and relations into the condensed paths. For the paths that cannot be further composed by rules, the heterogeneity between entities and relations prevents them from being immediately used to represent paths. Thus, the entities over paths are transformed into the semantic level representations similar to relations via a developed entity converter. Specifically, the entity converter contains two modules: attentive entity-to-type conversion for the KG with explicit types and entity-to-relation space projection for the KG without any type. We further employ the path encoder to learn the representations of bi-directional path sequences. Finally, the fusion representation learning of multiple paths is considered for learning entity and relation embeddings. The critical contributions of our work are summarized as follows:

\begin{itemize}
\item We put forward a novel KG inference method where semantics and data-driven path representation can be considered simultaneously to better learning KG embeddings.

\item Our model introduces horn rules to compose paths accurately for path representations. Moreover, to reduce the heterogeneity between entities and relations in the KG, two categories of entity converter strategies are proposed to transform the entities into the representations semantically similar to relations, which can work for any KG with or without explicit types. These two aspects are beneficial to balance the explainability and generalization of the model.

\item We conduct extensive experiments on three real-world KG datasets and compare our model's performance against several state-of-the-art methods. The experimental results demonstrate that our model consistently outperforms baselines on MRR and Hits@1 on both link prediction and path query answering task.

\end{itemize}

The remainder of this paper is organized as follows. We review related research in this area in Section~\ref{sec:related-work}. Section~\ref{sec:approach} provides the details of the proposed model. In Section~\ref{sec:experiments}, we conduct an extensive experimental evaluation and provide an analysis of the effectiveness of our model. Finally, we conclude the paper and discuss future research directions in Section~\ref{sec:conclusion}.

\section{Related Work}
\label{sec:related-work}

This section briefly describes prior works, including knowledge graph embedding approaches and path-based approaches for inference on knowledge graphs that are most related to our work.

\subsection{Knowledge Graph Embedding Approaches}
On account of the knowledge graph's symbolic nature, logical rules can be employed for KG inference. Many rule mining tools such as AMIE+ \cite{Galarrage:AMIE} and RLvLR \cite{RLvLR} extract logical rules from the KG, which could be leveraged to infer new triples. Although the KG inference approaches using logical rules have high accuracy, it cannot guarantee adequate generalization performance. In recent years, knowledge graph embedding approaches have been widely studied from a representation learning perspective, which could automatically explore the latent features from facts stored in the KG and learn the distributed representations for entities and relations. TransE ~\cite{Bordes:TransE} treats relations as translation operations between entity pairs to measure the compatibility of triples, which performs well on link prediction. TransH~\cite{Wang:TransH} extends TransE and projects an entity embedding into a relation-specific hyper-plane, endowing an entity with various representations. With a similar idea, TransR~\cite{Lin-a:TransR} projects an entity embedding into space concerning a relation. ComplEx~\cite{Trouillon:ComplEx} embeds entities and relations into complex vector space to handle the issue of inferring both symmetric and antisymmetric relations. ConvE~\cite{Dettmers:CNN} and ConvKB~\cite{ConvKB} leverage convolutional neural networks to capture the features of entities together with relations. An alternative direction is to focus on the matrix or tensor decomposition. KG2E ~\cite{he2015learning} models the uncertainties of data in a KG from the perspective of multivariate Gaussian distributions. He~\emph{et al.}~\cite{he2018knowledge} proposed a bayesian neural tensor decomposition approach to model the deep correlations or dependency between the latent factors in KG embeddings. Zhang~\emph{et al.}~\cite{zhang2020TKG} proposed a two-phase framework called TKGFrame for KG representation learning. However, these methods purely considering triples are unable to work in several complicated scenarios, such as path queries.

\subsection{Path-based Approaches for Inference on the Knowledge Graph}

In contrast to the approaches only relying on triples, the path-based reasoning models focus on capturing chains of reasoning expressed by paths from the KG. Early study on reasoning over paths is the Path Ranking Algorithm (PRA)~\cite{Lao:PRA}, which employs random walk on the KG and regards the paths as features to predict relations by the binary classifier. Some recent researches formulate path searching as sequential decision problems and aim to find the paths to infer the relations based on reinforcement learning (RL)~\cite{DeepPath,Das:MINERVA,Chen:DIVA}. DeepPath~\cite{DeepPath} is the first work that introduces reinforcement learning for searching paths between entity pairs. MINERVA \cite{Das:MINERVA} is an end-to-end system that develops an RL-based path finding procedure to reach the target entities for answering KG query. These multi-hop reasoning methods all spend much time in the complicated path finding process while obtaining limited inference performance.

Besides, some path and embedding-based approaches with higher efficiency and better performance are brought into focus. Particularly, pre-selected paths are generated by random walk for these methods. Neelakantan~\emph{et al.}~\cite{PathRNN} proposed a compositional vector space model with RNN to model relational paths on KGs. Guu \emph{et al.}~\cite{Guu:Traversing} modeled additive and multiplicative interactions between relation matrices along the path. PTransE \cite{Lin-b:PTransE} extends TransE by representing relational paths for learning entity and relation embeddings, and designs a path-constraint resource allocation (PCRA) algorithm to measure the reliability of relational paths. PaSKoGE~\cite{KBS:path-specific} minimizes a path-specific margin-based loss function while adaptively determining its margin. RPJE \cite{RPJE} introduces horn rules to compose paths with explainability for compositional representation learning on the KG. But these models ignore the entities in paths and only utilize the relations to represent paths. Both Jiang \emph{et al.}~\cite{Jiang2017AttentivePC} and Das \emph{et al.}~\cite{Das:Chain} developed the similar strategy to average the representations of entity types for each entity incorporated with relations to represent paths, neglecting that different entity types may play different roles in relation inference and these models cannot work on the KG without entity types.

\begin{figure*}
  \centering
  \includegraphics[scale=0.67]{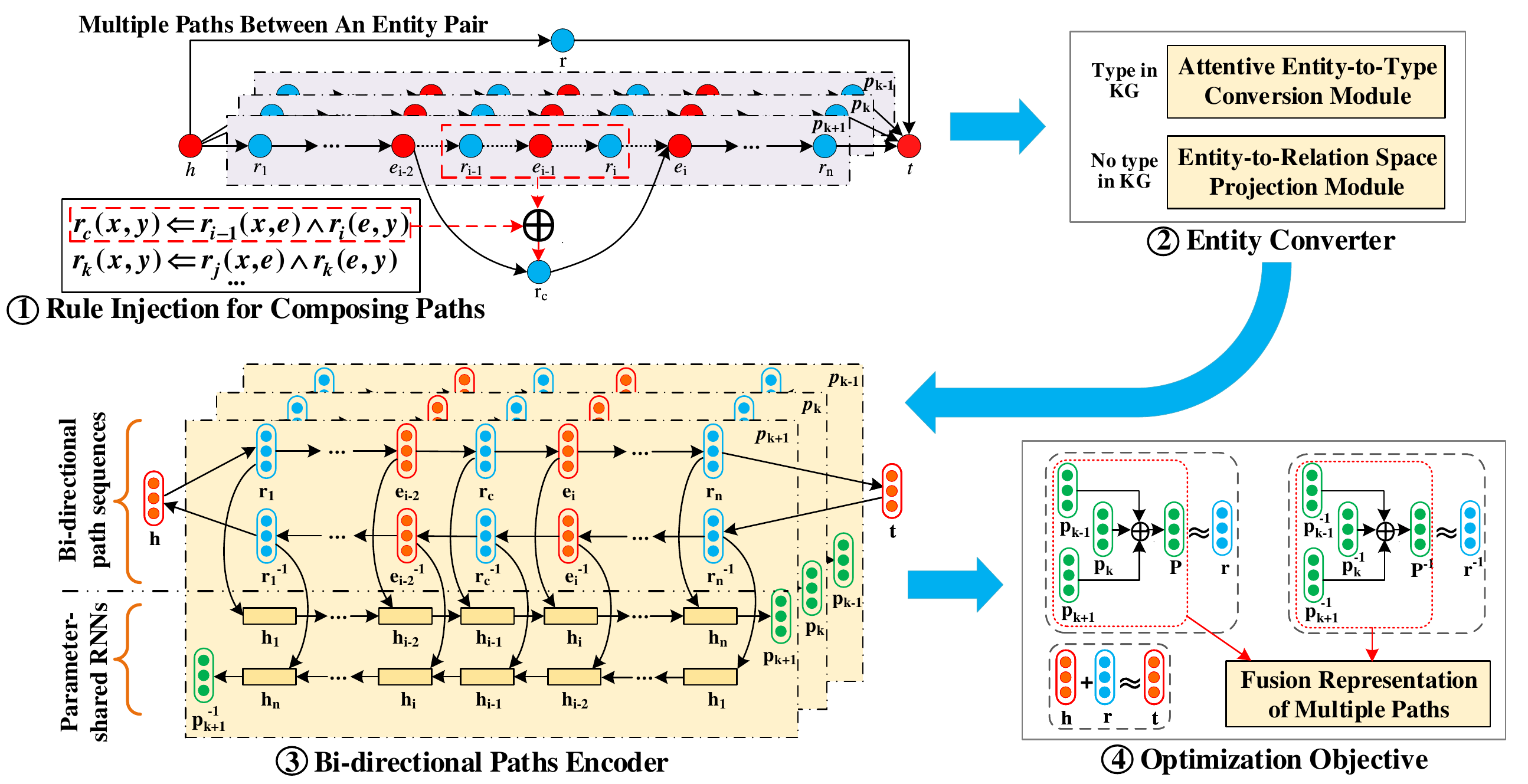}
  \caption{An overview of our model. The red units denote the entities, the blue units represent the relations, and the green units denote the path representations. The three paths shown in the figure are just for illustration. The other paths are processed in the same way as in the figure. Besides, a path between an entity pair could be transformed into bi-directional paths representations. In the optimization objective, the weight of each path $\alpha_i$ determines the extent to which this path contributes to the fusion representation learning of multiple paths.}
  \label{figure2}
\end{figure*}

\section{The Proposed Approach}
\label{sec:approach}
In this section, we extend the PCRA algorithm, which only extracts the relational paths in \cite{Lin-b:PTransE} by extracting the paths consisting of both entities and relations together with the normalized weight of each path. The pipeline of the overall architecture of our model is sketched in Fig.~\ref{figure2}. We first inject horn rules to generate the condensed paths $(\S \ref{section3.3})$. Then, on account of the paths whose length exceeds 1, the entities in paths are transformed into the representations semantically similar to relations via a general entity converter containing two modules for the KG both with and without types $(\S\ref{section3.4})$. Furthermore, a bi-directional paths encoder is developed to learn the path representations with the input of bi-directional path sequences between each entity pair $(\S\ref{section3.5})$. Finally, the optimization objective is proposed for learning embeddings of entities and relations by introducing the fusion representation learning of multiple paths $(\S\ref{section3.6})$.

\subsection{Rule Injection for Composing Paths}
\label{section3.3}

\begin{figure*}
\centering
\subfigure[A special scenario that indicates a path can be composed into a single relation by horn rules.]
{\includegraphics[width=0.42\textwidth]
{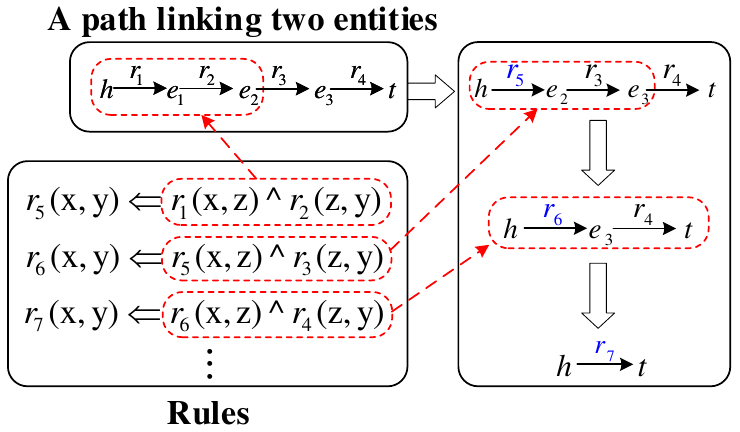}}
\hspace{0.10cm}
\subfigure[A general scenario that only a segment of the path can be composed by rules.] {\includegraphics[width=0.42\textwidth]
{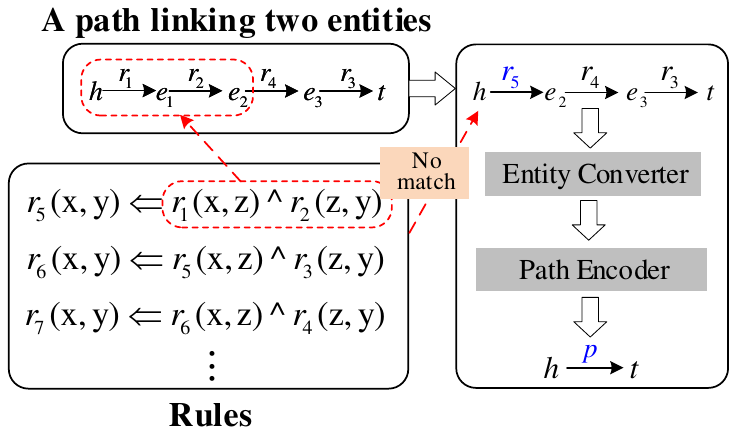}}
\caption{Two scenarios that are using horn rules for composing paths. Each rule is attached with a confidence such as $c_{1}, c_{2}, c_{3}$, and the ``Confi'' means the overall confidence of generating a condensed path by the rules. Note that the paths in the two scenarios are different.}
\label{figure3}
\end{figure*}

A horn rule is of the form \emph{head} $\Leftarrow$ \emph{body} (\emph{confi}), where \emph{body} is a conjunction of atoms $r_{i} (x_{i}, y_{i})$, e.g., $Nationality(x_{i},$ $y_{i})$, $x_{i}, y_{i}$ are variables that can be substituted with entities, \emph{head} is a single atom containing a specific relation, and \emph{confi} is a confidence value of the rule. In particular, the horn clause rules together with their confidence can be mined from KGs by employing some open source rule mining tool such as AMIE+ \cite{Galarrage:AMIE} and AnyBurl \cite{AnyBurl}. Among them, the rules of length 2 in the form of $r_{3}(x, y) \Leftarrow r_{1}(x, z) \wedge r_{2}(z, y)$ are appropriate to compose a path into a single relation or a condensed path in a high accuracy with explainability. Two scenarios of composing paths by rules are depicted in Fig.\ref{figure3}.


Theoretically, the horn rules of length 2 can compose any path greater than 1 in length, as illustrated in Fig.\ref{figure3} (a). In terms of a path linking two entities $h \xrightarrow{r_1} e_{1} \xrightarrow{r_2} e_{2} \xrightarrow{r_3} e_{3} \xrightarrow{r_4} t$, our model can start from the head entity and successively select a path segment by a 2-step path sliding window to search for a horn rule that matches this path segment. Then, the current segment of path can be composed into the relation the same as that in the rule head. Finally, our model loops the above path composition operation and the whole path can be condensed into a single relation whose embedding is the path representation. Meanwhile, the overall confidence is the multiplication of all the confidence of rules utilized in composing the path sequentially. However, it is hard to ensure every path segment can match a horn rule exactly right. An instance illustrated in Fig.\ref{figure3} (b) is the more general scenario, after composing a path segment once, none of the remaining path segments can be processed by rules. Therefore, we apply the following designed modules including the entity converter $(\S\ref{section3.5})$ and the path encoder $(\S\ref{section3.6})$ to produce the path representation. Therefore, it is necessary to develop a joint semantics and data-driven strategy for balancing the accuracy from rules and the generalization from path encoder.

\subsection{Entity Converter}
\label{section3.4}

Because both entities and relations are exploited for representing paths, the heterogeneity between entities and relations should be addressed. Therefore, we propose two kinds of entity converter mechanisms applicable to any KG.

\subsubsection{Attentive Entity-to-Type Conversion}

\begin{figure*}
  \centering
  \includegraphics[scale=0.50]{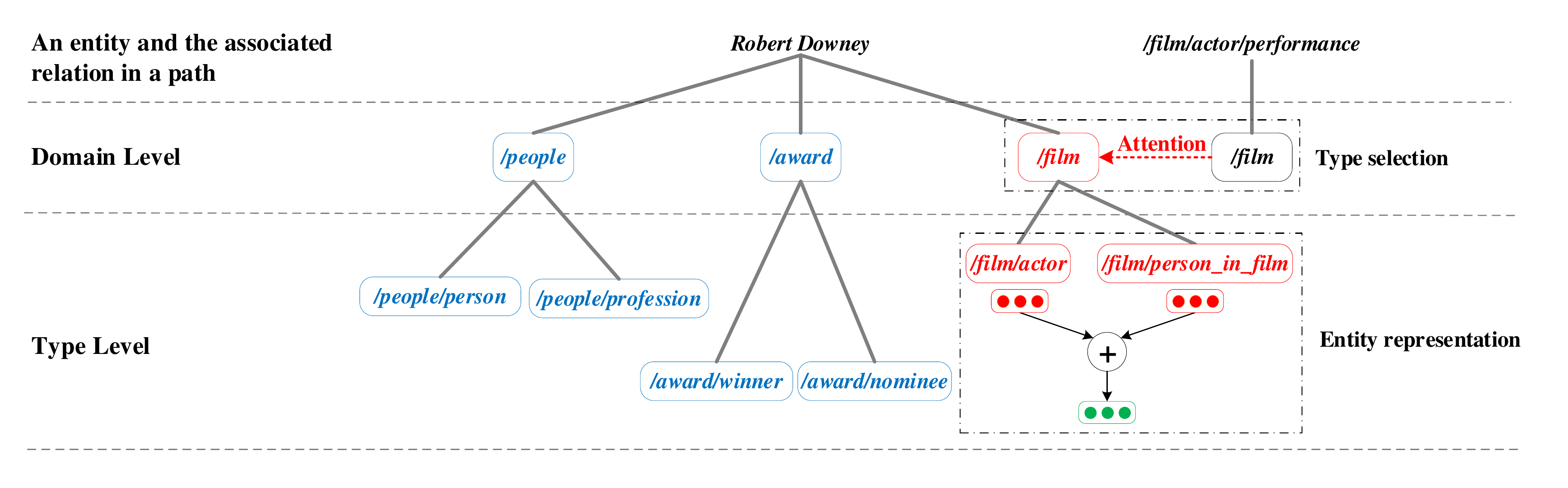}
  \caption{An example of attentive entity-to-type conversion in Freebase: Entity types with the hierarchical properties are viewed as a tree, where the entity is the root of the tree, and the depth of an entity type can model various semantic levels of hierarchy. The red symbols represent the entity's attended types by type-level attention mechanism, and the blue ones denote the other types of this entity. The green symbol stands for the composed type embedding, which could represent the entity in the path.}
  \label{figure4}
\end{figure*}

In some KGs such as Freebase, the entity types are always represented as type hierarchies following abstraction levels, where ``domain'' denotes a more abstract type and ``type'' represents a more specific type. The attentive entity-to-type conversion procedure focuses on the entity types that are most semantically relevant to the current path based on the type hierarchy and represent this entity with the selected types. An instance of representing entities by the associated types on Freebase is shown in Fig. \ref{figure4}. It is obvious that the type hierarchies of entity \small{$Robert\ Downey$} \normalsize can be represented as a tree structure. Specifically, an entity has multiple types distributed in both domain level and type level. Several types usually belong to the same domain, such as \small$/people/person$ \normalsize and \small$/people/profession$ \normalsize belong to \small$/people$\normalsize. Inspired by the fact that each relation corresponds to a unique domain included in the domains of entities, such as the relation \small$/film/actor$ $/performance$ \normalsize is associated with the domain \small$/film$\normalsize, we develop type-level attention to select the entity types which are closer to the context of the path. The domain \small$/film$ \normalsize of the entity is selected for the same domain as that of the relation, and the types \small$/film/actor$ \normalsize and \small$/film/$ $person\_in\_film$ \normalsize belonging to the domain \small$/film$ \normalsize could reflect the semantic categories of the entity \small$Robert\ Downey$ \normalsize more precisely and concretely.

Therefore, we develop an attentive entity-to-type conversion module for selecting the valuable types to represent an entity $e$ over a path with the following two steps: (1) Based on the designed type-level attention mechanism and the relation $r$ immediately after the entity $e$, the domain of the relation $r$ is denoted as $D(r)$. Then, among all the domains of the entity $e$, the same domain as $D(r)$ is selected in terms of the contextual semantics of the entity $e$ over the path. (2) We can represent the entity $e$ by adding all the embeddings of the types belonging to the selected domain $D(r)$. This attentive entity-to-type module guarantees both discrimination and generalization for representing the entity by entity types, which benefits from the semantic hierarchy of entity types. Furthermore, in allusion to an entity, $e$ with the associated relation $r$ in a path, the entity representation $\textbf{e}_{r}$ achieved by the attentive entity-to-type module can be formulated as:
\begin{flalign}\label{1}
\setlength{\mathindent}{0pt}
&\textbf{e}_{r} = \sum_{e_{t} \in T(e, r)} \textbf{e}_{t},\ T(e, r) = DT(attention(D(e), D(r)))&
\end{flalign}
where $\textbf{e}_{t}$ denotes an attended type embedding of entity $e$. $D(e)$ represents all the domains of an entity, and $D(r)$ denotes the domain belonging to a relation. $attention(D(e), D(r))$ is defined as the type-level attention operation for selecting the domain in the range of $D(e)$, which closely matches the contextual semantics of the current path. $DT(\cdot)$ could output the types corresponding to the input of the selected domain. $T(e, r)$ is denoted as the set of the types belonging to the selected domain of entity $e$ because of the semantic association with relation $r$.

\subsubsection{Entity-to-Relation Space Projection}
In allusion to the KG without entity types, we develop an entity-to-relation space projection module to project the entities along the paths into the corresponding relation space for reducing the heterogeneity between entities and relations. The representation of the entity $e$ projected into the space of relation $r$ is denoted as:
\begin{flalign}
\setlength{\mathindent}{0pt}
  &\textbf{e}_{r} = \textbf{P}_{r} \textbf{e}&
\label{eq2}
\end{flalign}
where $\textbf{e}_{r} \in \mathbb{R}^{k}$ denotes the projected representation of $e$ in the space of relation $r$, and $\textbf{e} \in \mathbb{R}^{k}$ is the embedding of entity $e$. $\textbf{P}_{r} \in \mathbb{R}^{k\times k}$ is defined as the projection matrix for projecting the entity $e$ into the space associated with relation $r$. Note that the quantity of $\textbf{P}_{r}$ is consistent with the number of relation in the KG.

\subsection{Path Encoder}
\label{section3.5}
We introduce a path encoder with the input of bi-directional paths and a shared parameter architecture of RNN, enabling fewer parameters to output any relation embedding for representing the path. We first define the path sequence as the input of the path encoder as follows:
\begin{flalign}
\setlength{\mathindent}{0pt}
  & \textbf{p} = \textbf{r}_{1}, \textbf{e}_{r1}, \textbf{r}_{2}, \textbf{e}_{r2}, \cdots, \textbf{r}_{n} &
\label{eq3}
\end{flalign}
where $n$ denotes the number of relations in the path sequence, and then the length of the path sequence should be $2n-1$. The relation embeddings $\textbf{r}_{i}, i = 1, \cdots, n$ and the entity representations $\textbf{e}_{r_{i}}, i = 1, \cdots, n - 1$ derived from the entity converter are the components of a path sequence.

To predict a relation which closely represents a path by consuming the path sequence, the path encoder is defined as:
\begin{flalign}
\setlength{\mathindent}{0pt}
  & \overrightarrow{h}_{t} = ReLU(W_{h} \overrightarrow{h}_{t-1} + W_{i} \overrightarrow{x}_{t}) &
\label{eq4}
\end{flalign}
in which $\overrightarrow{h}_{t}\in\mathbb{R}^{n}$ and $\overrightarrow{h}_{t-1}\in\mathbb{R}^{n}$ are hidden states of the RNN at step t and t-1, respectively. $W_{h} \in \mathbb{R}^{n\times n}$ and $W_{i}\in \mathbb{R}^{n\times k}$ denote the parameter matrices of the RNN. $\overrightarrow{x}_{t} \in \mathbb{R}^{n}$ is the representation vector of the path component at step $t$. Specifically, when $t$ is an odd number, $\overrightarrow{x}_{t}$ is the $t$-th relation representation in the path sequence, and when $t$ is an even number, $\overrightarrow{x}_{t}$ is the $(t/2)$-th entity representation in the path sequence. ReLU is the rectifier linear unit and we select it for its best performance compared to other commonly used activation functions. In Eq.\ref{eq4}, the final hidden state of the path encoder is output as the predicted relation embedding and also the path representation. In particular, for a path of length 1 containing only one relation, the path encoder directly outputs the embedding of this relation.

Significantly, according to the added inverse relations, each original path extracted from the KG can be converted into the bi-directional path sequences. We can achieve both forward and inverse path representations with the parameter-shared path encoder for bi-directional paths linking entities.

\subsection{Objective Formalization}
\label{section3.6}
The actual human-like inference mechanism infers according to the multiple paths organized by a graph structure. Therefore, to comprehensively exploit the semantics of multiple paths between entities, we propose the fusion representation learning of multiple paths to measure the correlation between a relation with multiple paths in the same direction, and the weight of each path determines the extent to which this path contributes to the fusion representation learning of multiple paths. Two energy functions concerning triples and multiple paths are developed as follows:
\begin{flalign}
\setlength{\mathindent}{0pt}
  &E_{1}(h, r, t) = \Vert \textbf{h} + \textbf{r} - \textbf{t}\Vert_{L_{1}/L_{2}} \label{eq5a}&\\
  &E_{2}(r, \mathcal{P}) = \frac{1}{\sum{\alpha_{i}}}c_{i}\alpha_{i}\sum_{p_{i}\in \mathcal{P}}\Vert \textbf{r} - RNN(p_{i}) \Vert_{L_{1}/L_{2}}&
\label{eq5}
\end{flalign}
where $E_{1}$ captures the translational principle of each triple $(h, r, t)$ and is the primary function depending on single triples. $\textbf{h}, \textbf{r}, \textbf{t}$ denote the embeddings of head entity, relation and tail entity, respectively. $E_{2}$ models the dissimilarity between multiple paths and the direct relationship from head entity to tail entity. Here $\mathcal{P}$ is a set of paths in the same direction as relation $r$, and $p_i$ is one of the paths in $\mathcal{P}$. $\alpha_{i}$ is the weight of path $p_{i}$ obtained in the process of path extraction by our extended PCRA algorithm. $c_{i}$ is the overall confidence of composing path $p_{i}$ by horn rules mentioned in section~\ref{section3.3}. $RNN(p_{i})$ is the representation of path $p_{i}$ via the path encoder defined in Eq.\ref{eq4}.

Based on the energy functions stated in Eqs.\ref{eq5a}, \ref{eq5} and the negative sampling strategy, we propose a two-component loss function $L$ comprised of $L_{1}$ and $L_{2}$ for training our model:
\begin{flalign}
\setlength{\mathindent}{0pt}
  & L = \sum_{(h, r, t)\in \mathcal{T}} (L_{1} + \lambda L_{2})\label{eq6}&\\
  & L_{1} = \sum_{(h^{'}, r^{'}, t^{'}) \in \mathcal{T}^{'}} \big[\gamma_{1} + E_{1} (h, r, t) - E_{1}(h^{'}, r^{'}, t^{'})\big]_{+}\label{eq7}&\\
  &L_{2} = \sum_{r^{'} \in \mathcal{T}^{'}} \left[\gamma_{2} + \frac{E_{2} (r, \mathcal{P}) + E_{2}(r^{-1}, \mathcal{P}^{-1})}{2} - E_{2}(r^{'}, \mathcal{P})\right]_{+}\label{eq8}&
\end{flalign}
in which $L_{1}$ is the triple-specific loss and $L_{2}$ is the path-specific loss, traded off by a parameter $\lambda$. $\gamma_{1}$ and $\gamma_{2}$ are two margins in $L_{1}$ and $L_{2}$, respectively. $[x]_{+}$ returns the maximum value between $x$ and 0. $r^{-1}$ represents the inverse version of the relation $r$ and $\mathcal{P}^{-1}$ denotes the set of path sequences in the inverse direction of $\mathcal{P}$. $\mathcal{T}$ is the set of positive triples observed in the KG. We define the negative sample set of $\mathcal{T}$ as
\begin{flalign}
\setlength{\mathindent}{0pt}
  &\mathcal{T}^{'} = {(h^{'}, r, t)\cup(h, r^{'}, t)\cup(h, r, t^{'})}&
\label{eq9}
\end{flalign}
where $\mathcal{T}^{'}$ is comprised of the negative triples which are generated by randomly replacing one of the components in positive triples by another one and wiping out the generated triples already in $\mathcal{T}$.

\pheadWithSpace{Algorithm}
The complete training procedure of our proposed model is shown in Algorithm~\ref{alg:kgInfer}. We take $\mathcal{G}$, $\mathcal{A}$, $\gamma_{1}, \gamma_{2}$, $\lambda$, $T$, $b$ as input. Before invoking the main algorithmic loop, we initialize relevant variables and set the number of epochs as the stop criterion for the loop. The algorithm starts updating corresponding entity and relation embeddings, and variables recurrently until the convergence or stop criterion is met. Finally, the completed entity and relation embeddings are generated after several epochs.

\begin{algorithm}
\caption{Training framework of our proposed model}
\label{alg:kgInfer}
\LinesNumbered
\KwIn{
$\mathcal{G}$: Training set \newline
$\mathcal{A}$: The set of all the paths extracted from $\mathcal{G}$ \newline
$\gamma_{1}$, $\gamma_{2}$: Margins \newline
$\lambda$: The weight for trade-off \newline
$T$: The total number of epochs \newline
$b$: The batch size
}
\textbf{Initialize} Entity embeddings $\textbf{e}$, relation embeddings $\textbf{r}$, the projection matrix $\textbf{P}_{r}$, the weights of RNN $\textbf{W}_{h}, \textbf{W}_{i}$ randomly

\For{epoch = 1, 2, \dots, T}{
    $\mathcal{T}_{batch}$ $\leftarrow$ Sample($\mathcal{G}$, b) \tcc{Sample a minibatch of triples}
    $\mathcal{S}_{batch}$ $\leftarrow$ $\emptyset$ \tcc{Initialize the set of input instances}
    \For{$(h, r, t)\in \mathcal{T}_{batch}$}{
        $(h^{'}, r^{'}, t^{'})$ $\leftarrow$ NegativeSample($\mathcal{G}$, $(h, r, t)$) \tcc{Sample a corrupted triple}
        $\mathcal{P}$ $\leftarrow$ Generate($\mathcal{A}$, $(h, r, t)$) \tcc{Get a set of paths between $(h, t)$}
        \For{$p \in \mathcal{P}$}{
        \If {a Horn rule $R$ can be matched with $p$}{
            $p^{'}$ $\leftarrow$ Compose($p$, $R$)  \tcc{Compose $p$ with the rule $R$}
            $p$ $\leftarrow$ Replace($p$, $p^{'}$) \tcc{Replace $p$ with $p^{'}$}
        }
        \eIf{Entity types in the dataset}{
            Converting entities in path p according to Eq.\ref{1}
        }{
            Converting entities in path p according to Eq.\ref{eq2}
        }
            $\textbf{p}$ $\leftarrow$ Encode($p$) \tcc{Encoding path $p$ by Eq.\ref{eq4}}
        }
        $\mathcal{P}^{-1}$ $\leftarrow$ Reverse($\mathcal{P}$) \tcc{Get the set of inverse paths}
        \For{$p \in \mathcal{P}^{-1}$}{
        The same procedure as in line 7-16
        }
        $\mathcal{S}_{batch}$ $\leftarrow$ $\mathcal{S}_{batch}$ $\cup\{(h, r, t),(h^{'}, r^{'}, t^{'}),\mathcal{P},\mathcal{P}^{-1}\}$
    }
    Update embeddings $\textbf{e}$, $\textbf{r}$, and parameters $\textbf{P}_{r}$, $\textbf{W}_{h}, \textbf{W}_{i}$ w.r.t. \newline $\sum_{\left((h, r, t),(h^{'}, r^{'}, t^{'}),\mathcal{P},\mathcal{P}^{-1}\right) \in \mathcal{S}_{batch}} \nabla \big[\gamma_{1} + E_{1}(h, r, t)$ $- E_{1}(h^{'}, r^{'}, t^{'})\big]_{+} + \lambda \big[\gamma_{2} + \frac{E_{2} (r, \mathcal{P}) + E_{2}(r^{-1},\mathcal{P}^{-1})}{2}$ $- E_{2}(r^{'}, \mathcal{P})\big]_{+}$
}
\end{algorithm}

\section{Experiments}
\label{sec:experiments}
In this section, we will describe the datasets, and the experimental setup applied in our experiments. In specific, We compared several tools like AMIE+ \cite{Galarrage:AMIE} and AnyBurl \cite{AnyBurl}, and AMIE+ is selected for its convenience and good performance of mining more high-quality (confidence > 0.7) horn rules for path composition. We empirically evaluate our proposed approach with state-of-the-art baselines on two tasks, including link prediction and path query answering.

\subsection{Datasets and Experimental Setup}

\subsubsection{Datasets}
We employ three widely-used benchmark datasets for the experiments, including FB15K~\cite{Bordes:TransE}, WN18~\cite{Bordes:TransE}, and NELL-995~\cite{Mitchell:nell}. The statistics of these datasets are exhibited in Table~\ref{table1}. In particular, FB15K contains 90 domains and 3,853 types of entities in total, which can be conducted by both attentive entity-to-type conversion and entity-to-relation space projection modules. In contrast, only entity-to-relation space projection module can be evaluated on WN18 and NELL-995 due to no types in these two datasets.

 \begin{table*}[width=2.2\linewidth,cols=4,pos=h]
 \caption{Statistics of the experimental datasets.}
 \centering
 \renewcommand\tabcolsep{8.6pt}
 \begin{tabular}{ccccccc}
 \toprule
Dataset		& \#Relation	& \#Entity	& \#Train	& \#Valid	& \#Test	& \#Rule\\
 \midrule
 FB15K		   & 1,345		 & 14,951   & 483,142	& 50,000	& 59,071	   & 899\\
 WN18        & 18        & 40,943    & 141,442   & 5,000     & 5,000    & 77\\
 NELL-995    & 200       & 75,492    & 123,370    & 15,000     & 15,838    &201\\
 \bottomrule
 \end{tabular}
 \label{table1}
 \end{table*}

\subsubsection{Hyperparameter settings}

All the experiments are performed on an Intel i9-9900X CPU with 64 GB main memory. We initialize all the baselines with the parameter settings in the corresponding papers and then turn them on experimental datasets for best performance for a fair comparison. For our model, we train it by setting batch size of 1024 on each dataset, with grid search for selecting the best hyperparameters. We select the embedding dimension of both entity and relation $k \in \{50, 100, 150,$ $200\}$, the learning rate $\delta \in \{0.001, 0.005,$ $0.01, 0.02, 0.05\}$, the margins $\gamma_{1}, \gamma_{2} \in \{0.5,$ $1.0, 1.5, 2.0, 3.0, 5.0\}$, and the weight for trade-off $\lambda \in \{0.5,$ $1.0, 1.5, 2.0, 3.0\}$. During the training phrase, we select $L_{1}$ norm for all the equations containing norms and apply $L_{1}$ regularizer for all the parameters. The optimal configurations are obtained by trial and error: \{$k$ = 100, $\delta$ = 0.001, $\gamma_{1}$ = $\gamma_{2}$ = 1.0, $\lambda$ = 1.0\} on FB15K and NELL-995 datasets, \{$k$ = 50, $\delta$ = 0.01, $\gamma_{1}$ = 1.0,\ $\gamma_{2}$ = 2.0, $\lambda$ = 0.5\} on WN18 dataset. We also perform grid search on the maximum-length of paths 2 and 3, and the length of paths is set no longer than 2 due to the comprehensive consideration of the training efficiency and inference performance.

\subsubsection{Baseline Methods}
For comparison, we select several typical and competitive baselines, which are categorized into two groups: (1) the models that utilize triples alone, including TransE~\cite{Bordes:TransE}, TransH~\cite{Wang:TransH}, TransR~\cite{Lin-a:TransR}, HolE~\cite{HolE}, ComplEx~\cite{Trouillon:ComplEx} and ConvE~\cite{Dettmers:CNN}. (2) The models that introduce paths from the KG, including embedding-based PTransE~\cite{Lin-b:PTransE} and reinforcement learning-based MINERVA~\cite{Das:MINERVA}. Our model falls into the second group as it leverages both triples and paths.

\subsection{Link Prediction}
\label{section 4.2}
Link prediction is a classical evaluation task that tries to predict the missing head or tail entity for an incomplete triple $(h, r, t)$. This task could be regarded as a simple question-answering in a sense.

\subsubsection{Evaluation Protocol}
For evaluation, we follow the standard protocol: for each triple $(h, r, t)$ in the test set, we replace the head or tail entity with all entities in the KG to create a set of candidate triples. Then, The candidate triples can be ranked in ascending order according to the scores calculated by
\begin{flalign}
\setlength{\mathindent}{0pt}
  & E_{r}(h, r, t, \mathcal{P}) = E_{1}(h, r, t) + \lambda \frac{(E_{2}(r, \mathcal{P}) + E_{2}(r^{-1}, \mathcal{P}^{-1}))}{2}&
\label{eq11}
\end{flalign}
Then, we obtain the rank of the correct triple and report the performance using two standard metrics: the mean reciprocal rank (MRR) and the proportion of ranks no larger than n (Hits@n). Here, higher MRR or Hits@n accounts for better performance. Besides, all evaluation results are shown under the setting ``Filter'', i.e., removing any candidate triple already exists in the training, validation, or test set except for the correct one.

\subsubsection{Experimental Results}
We conduct the link prediction experiments for the baselines TransE, TransH, TransR $\footnote{The code for TransE, TransR, and TransH is from \url{https://github.com/thunlp/Fast-TransX}.}$ and HolE $\footnote{\url{https://github.com/thunlp/OpenKE}}$ on NELL-995 while PTransE $\footnote{\url{https://github.com/thunlp/KB2E}}$ and MINERVA $\footnote{\url{https://github.com/ashiqueh/MINERVA}}$ on all the three datasets by their source codes. The other results of baselines are achieved from the corresponding original papers. As for our model, two entity converter modules attentive entity-to-type conversion and entity-to-relation space projection are abbreviated as EC1 and EC2, respectively.

\begin{table*}[width=2.07\linewidth]
 \caption{Evaluation results of comparison models on FB15K, WN18 and NELL-995. The best and second best results in each column is boldfaced and underlined, respectively. Note that both Entity-to-Type and Entity-to-Relation modules are available for FB15K.}
 \begin{tabular}{c|c|c|c|c|c|c|c|c|c|c|c|c}
 \toprule
     & \multicolumn{4}{c}{FB15K}    & \multicolumn{4}{|c|}{WN18}	& \multicolumn{4}{c}{NELL-995}\\
     \cline{2-5} \cline{6-9} \cline{10-13}
     Model                 & {}        & \multicolumn{3}{c|}{Hits@n}   & {} & \multicolumn{3}{c|}{Hits@n}    & {} &\multicolumn{3}{c}{Hits@n}\\
     \cline{3-5} \cline{7-9} \cline{11-13}
      	        	            & MRR       & 1         & 3         & 10	    & MRR    & 1      & 3        & 10     & MRR   & 1         & 3     & 10\\
 \midrule
 TransE     & 0.378		& 0.229     & 0.468     & 0.635     & 0.452  & 0.087  & 0.819    & 0.928  & 0.219  & 0.150 & 0.247  & 0.352\\
 TransH     & 0.280		& 0.171     & 0.329     & 0.641     & 0.358  & 0.045  & 0.618    & 0.814  & 0.223     & 0.152    & 0.250  & 0.358\\
 TransR     & 0.343		& 0.217     & 0.405     & 0.583     & 0.603  & 0.334  & 0.875    & 0.942  & 0.232    & 0.262    & 0.309   & 0.382\\
 HolE       & 0.524	  & 0.403     & 0.616     & 0.727     & 0.938  & 0.930    & 0.943   & 0.946     & 0.299     & 0.255   & 0.304  & 0.386\\
 ComplEx    & \underline{0.692}	  & 0.599     & \underline{0.759}     & 0.839     & 0.941  & \underline{0.936}    & \underline{0.945}   & 0.947     & -     & -   & -  & -\\
 ConvE      & 0.657		& 0.558     & 0.723     & 0.831     & \underline{0.943}  & 0.935  & \textbf{0.946}    & \textbf{0.956}  & -     & -   & -  & -\\
 \midrule
 PTransE-ADD    & 0.679		& 0.565     & \textbf{0.768}     & 0.855     & 0.890  & 0.931     & 0.942    & 0.945 & 0.304   & 0.234   & 0.337  & \underline{0.437}\\
 PTransE-RNN    & 0.539		& 0.376     & 0.636     & 0.822     & 0.886  & 0.851     & 0.919    & 0.939 & 0.286
 & 0.232   & \underline{0.386}  & 0.423\\
 MINERVA       & 0.387		& 0.321     & 0.426     & 0.511     & 0.866  & 0.827     & 0.899    & 0.926 & \underline{0.328}     & \underline{0.274}   & 0.359  & 0.433\\
 \midrule
 Ours (EC1)	        & \textbf{0.716}	& \textbf{0.701}     & 0.756  & \textbf{0.868}   & -	& -  & -  & -   & -	& -  & -  & -\\
 Ours (EC2)	    & \textbf{0.715}	& \underline{0.651}     & 0.750  & \underline{0.857}   & \textbf{0.946}	& \textbf{0.940}     & \underline{0.945}  & \underline{0.952} & \textbf{0.350} & \textbf{0.282}  & \textbf{0.402}  & \textbf{0.475}\\
 \bottomrule
 \end{tabular}
 \label{table2}
 \end{table*}

The evaluation results on FB15K, WN18 and NELL-995 are summarized in Table \ref{table2}. It indicates that our approach outperforms the state-of-the-art models in most cases. Some further analyses are listed as follows:

\begin{itemize}
  \item Our model consistently outperforms other baselines in terms of MRR and Hits@1 on all the three datasets. Compared to the best baselines ComplEx and PTransE-add, our model obtains the performance gains as: (1) Compared with ComplEx: 3.3\%/17.0\% of MRR/Hits@1 on FB15K, 0.5\%/0.4\% of MRR/Hits@1 on WN18. (2) Compared with PTransE-add: 5.5\%/24.1\% of MRR/ Hits@1 on FB15K, 6.3\%/1.0\% of MRR/Hits@1 on WN18, 15.1\%/20.5\% of MRR/Hits@1 on NELL-995. These results emphasize the superiority of our approach in high-precision link prediction.

  \item Specific to the results on FB15K, our model with attentive entity-to-type conversion performs similarly but slightly better than that with entity-to-relation space projection. This result illustrates the effectiveness of the proposed entity converter mechanism on any KG, whether it contains entity types.

  \item All the path-based baselines and our model outperform the other models using triples alone on the more sparse dataset NELL-995, verifying the higher performance gains on link prediction benefit from exploiting multi-hop paths to create more semantic relationships between entities.

  \item Our approach outperforms other path-based baselines PTransE-ADD, PTransE-RNN and MINERVA mainly due to incorporating both entities and relations in path sequences as well as the joint semantics and data-driven path representation for introducing more effective path representations in KG inference.
\end{itemize}

\subsubsection{Ablation Study}
We implement the ablation study to analyze the effects of the core components by removing them from our full model and evaluating the performance on FB15K and WN18. As shown in Table \ref{table3}, compared to our whole model, the model that removes Entity Converter (-EC) drops 5.88\% of Hits@10 on FB15K, and the model removes Bi-directional Path (-BDP) drops 3.8\% of Hits@10 on FB15K, and the model that removes Rules (-Rule) drops 1.72\% of Hits@10 on FB15K, respectively. The ablation study emphasizes the significance of all the components leveraged in our model, especially the entity converter.

\begin{table}
\centering
\caption{Ablation study of our model on FB15K and WN18.}
\begin{tabular}{c|cc|cc}
\toprule
& \multicolumn{2}{c}{FB15K}    & \multicolumn{2}{|c}{WN18}\\
    Model                 & MRR   & Hits@10 & MRR   & Hits@10 \\
\midrule
Our full model	            & 0.715 	    & 0.868	& 0.946     & 0.952     \\
\midrule
-EC                         & 0.691       & 0.817 & 0.896     & 0.932       \\
-BDP                        & 0.709       & 0.835 & 0.925     & 0.943    \\
-Rule                       & 0.702       & 0.853 & 0.933     & 0.947 \\
\bottomrule
\end{tabular}
\label{table3}
\end{table}

\subsection{Path Query Answering}
To evaluate the performance of KG inference concentrating on paths, we conduct the complex multi-hop path query answering containing two sub-tasks: (1) A path query for entity contains an initial entity and a path sequence and answering this query is to predict the target entity by the initial entity and the path. For example, the path query $Iron\ Man$ $\xrightarrow{Actor} Robert\ Downey \xrightarrow{Lives\_in} ?$ means the multi-hop reasoning ``where does Robert Downey, the actor of Iron Man, live?''. (2) A path query for relation consists of an entity pair and multiple paths between the two entities. Answering this query is to infer the relation connecting the entity pair by both entities and multiple paths.

\begin{table*}[width=2.07\linewidth]
\caption{Evaluation results of entity prediction of path query answering.}
\renewcommand\tabcolsep{6pt}
\begin{tabular}{c|cccc|cccc|cccc}
\toprule
    & \multicolumn{4}{c}{FB15K}    & \multicolumn{4}{|c|}{WN18}	& \multicolumn{4}{c}{NELL-995}\\
    \cline{2-5} \cline{6-9} \cline{10-13}
    Model                 & {}  & {}       & \multicolumn{2}{c|}{Hits@n}   & {} & {} & \multicolumn{2}{c|}{Hits@n}    & {}& {}  &\multicolumn{2}{c}{Hits@n}\\
    \cline{4-5} \cline{8-9} \cline{12-13}
     	        	            & MR   & MRR    & 1         & 10	    & MR  & MRR  & 1      & 10     & MR  & MRR  & 1         & 10\\
\midrule
PTransE-ADD    & 38		& 0.576     & 0.433     & 0.862  & 20     & 0.938    & 0.893 & \textbf{0.993}   & 2068  & 0.686 & 0.625    & 0.812\\
PTransE-RNN   & 42		& 0.564     & 0.405     & 0.850   & 22  & 0.933     & 0.895    & 0.990
& 2166   & 0.681  & 0.607 & 0.816\\
\midrule
Ours (EC1)	        & \textbf{28}	& 0.750     & \textbf{0.756} & \textbf{0.883}   & -	& -  & -  & -   & -	& -  \\
Ours (EC2)	    & 30	& \textbf{0.751}     & \textbf{0.756} & 0.879   & \textbf{9} & \textbf{0.988}	& \textbf{0.987}     & 0.991 & \textbf{1934} & \textbf{0.715} & \textbf{0.657}  & \textbf{0.826}\\
\bottomrule
\end{tabular}
\label{table5}
\end{table*}

\begin{table*}[width=2.07\linewidth]
\caption{Evaluation results of relation inference of path query answering. We just set n = 1 of Hits@n to highlight the performance differences among the models.}
\renewcommand\tabcolsep{8pt}
\begin{tabular}{c|ccc|ccc|ccc}
\toprule
    & \multicolumn{3}{c}{FB15K}    & \multicolumn{3}{|c|}{WN18}	& \multicolumn{3}{c}{NELL-995}\\
    \cline{2-4} \cline{5-7} \cline{8-10}
    Model                           & MR        & MRR   & Hits@1         	    & MR  & MRR  & Hits@1           & MR  & MRR  & Hits@1         \\
\midrule
TransE          & 2.522     & 0.578      & 0.857    & 1.833     & 0.774   &  0.650   & 34.17     & 1.374 & 0.315\\
PTransE-ADD    & 1.407		& 0.934      & 0.942     & 1.063     & 0.970  & 0.995     & 1.683    & 1.858 & 0.887   \\
PTransE-RNN    & 1.435		& 0.917      & 0.938     & 1.041     & 0.983  & 0.997     & 1.728    & 1.854 & 0.882\\
\midrule
Ours (EC1)	        & \textbf{1.322} & \textbf{0.970}     & 0.971    & -	& -  & -  & -   & -	& -  \\
Ours (EC2)	    & 1.335     	& 0.967    & \textbf{0.975}     & \textbf{1.016} & \textbf{0.996} & \textbf{0.998}   & \textbf{1.625} & \textbf{1.863}	& \textbf{0.896} \\
\bottomrule
\end{tabular}
\label{table6}
\end{table*}

\begin{table*}[width=2.07\linewidth]
  \caption{Path query answering examples of relation inference with explainability on NELL-995.}
  \renewcommand\tabcolsep{10pt}
  \begin{tabular}{l|l}
  \toprule
    \multicolumn{2}{c}{Example \#1}\\
  \midrule
    Path query    & path: $kobe\_bryant \xrightarrow{\textbf{\textcolor{red}{athlete\_plays\_for\_team}}}	los\_angeles\_lakers \xrightarrow{\textbf{\textcolor{blue}{team\_plays\_in\_league}}} nba$\\
                  & query: $(kobe\_bryant, ?, nba)$ \\
  \midrule
    The matching rule     & $\textbf{athlete\_plays\_in\_league (x, y)} \Leftarrow \textbf{\textcolor{red}{athlete\_plays\_for\_team} (x, z)}$ \\
                          &\ \ \ \ \ \ \ \ \ \ \ \ \ \ \ \ \ \ \ \ \ \ \ \ \ \ \ \ \ \ \ \ \ \ \ \ \ \ \ \ \ \ \ $\wedge \textbf{\textcolor{blue}{~team\_plays\_in\_league}  (z, y)}$ (confidence: 0.8) \\
  \midrule
    Rank of the correct relation    & $\textbf{athlete\_plays\_in\_league}$\ \ \ Rank:\ \textbf{1} \\
  \midrule
    The other top 3 relations    & athlete\_plays\_sport\ \ \ \ Rank:\ 2 \\
                                           & coaches\_in\_league\ \ \ \ \ Rank:\ 3 \\
  \bottomrule
  \toprule
    \multicolumn{2}{c}{Example \#2}\\
  \midrule
    Path query    & path: $jonathan \xrightarrow{\textbf{\textcolor{red}{has\_sibling}}} simon \xrightarrow{\textbf{\textcolor{blue}{~person\_born\_in\_city}}} york$\\
                  & query: $(jonathan, ?, york)$ \\
  \midrule
    The matching rule     & $\textbf{person\_born\_in\_city(x, y)} \Leftarrow \textbf{\textcolor{red}{has\_sibling} (x, z)}$ \\
                          &\ \ \ \ \ \ \ \ \ \ \ \ \ \ \ \ \ \ \ \ \ \ \ \ \ \ \ \ \ \ \ \ \ \ \ \ \ \ $ \wedge \textbf{\textcolor{blue}{~person\_born\_in\_city} (z, y)}$ (confidence: 1.0) \\
  \midrule
    Rank of the correct relation    & $\textbf{person\_born\_in\_city}$\ \ \ Rank:\ \textbf{1} \\
    \midrule
    The other top 3 relations    & person\_moved\_to\_state\_or\_province\ \ \ \ \ Rank:\ 2 \\
                                 & person\_has\_citizenship\ \ \ \ \ \ \ \ \ \ \ \ \ \ \ \ \ \ \ \ \ \ \ \ \ Rank:\ 3 \\
  \bottomrule

      \toprule
    \multicolumn{2}{c}{Example \#3}\\
  \midrule
    Path query    & path: $bill \xrightarrow{\textbf{\textcolor{red}{top\_member\_of\_organization}}} microsoft\_corp \xrightarrow{\textbf{\textcolor{blue}{company\_also\_known\_as}}} microsoft$\\
                  & query: $(bill, ?, microsoft)$ \\
  \midrule
      The matching rule     & $\textbf{person\_leads\_organization (x, y)} \Leftarrow \textbf{\textcolor{red}{top\_member\_of\_organization} (x, z)}$ \\
                          &\ \ \ \ \ \ \ \ \ \ \ \ \ \ \ \ \ \ \ \ \ \ \ \ \ \ \ \ \ \ \ \ \ \ \ \ \ \ \ \ \ \ \ \ \ $ \wedge \textbf{\textcolor{blue}{~company\_also\_known\_as} (z, y)}$ (confidence: 0.83) \\
  \midrule
    Rank of the correct relation    & $\textbf{person\_leads\_organization}$\ \ \ Rank:\ \textbf{1} \\
  \midrule
    The other top 3 relations    & person\_has\_job\_position\ \ \ \ \ \ \ \ \ \ \ \ Rank:\ 2 \\
                                 & athlete\_belongs\_to\_organization\ \ \ Rank:\ 3 \\
  \bottomrule
  \end{tabular}
  \label{table7}
\end{table*}


\subsubsection{Evaluation Protocol}
To create the test sets, the paths for testing are extracted from the whole graph consisting of both training and test triples with our paths extraction algorithm, which is the same training procedure. The paths already exploited for training should be removed. Then, we evaluate the path query answering tasks on the test sets generated by both the original test triples and the corresponding multiple test paths. The triples of the constructed test sets on FB15K, WN18 and NELL-995 are 113,676/9,492/12,398, and the paths on three constructed test sets are 1,825,415/16,934/97,024, respectively.

On account of the path query answering for entity prediction, a query $h\rightarrow r_{1} \rightarrow e_{1} \cdots \rightarrow r_{l}\rightarrow ?$ is created by removing the tail entity of each test instance and select the path with the highest weight if there are multiple paths. We rank the correct answer with the other answers based on the scores in ascending order according to the score function as
\begin{flalign}
\setlength{\mathindent}{0pt}
    & E_{q} (h, p, t) = \Vert \textbf{h} + RNN(p) - \textbf{t}\Vert_{L_{1}/L_{2}}\nonumber\\
                        &\ \ \ \ \ \ \ \ \ \ \ \ \ \ \ \ \ \ \ \ + \Vert \textbf{t} + RNN(p^{-1}) - \textbf{h}\Vert_{L_{1}/L_{2}} &
\label{eq10}
\end{flalign}where the forward path $p$ and its inverse version $p^{-1}$ are considered jointly.

To evaluate the path query answering for relation inference, we calculate the score via the energy function defined in Eq.~\ref{eq11} and rank the correct answers. For evaluation on both path query answering sub-tasks, we employ the ``Filter'' setting to report the mean rank (MR) and the mean reciprocal rank (MRR) over all the test cases, and also the proportion of test cases obtaining the correct answer in the top n (Hits@n).

Among all the baselines utilized in link prediction, only PTransE \cite{Lin-b:PTransE} is applicable for both path query answering sub-tasks because the other baselines, including the methods that only consider triples and the RL-based path reasoning approaches, are unable to infer with the given paths. Specific to path query answering for relation inference, TransE \cite{Bordes:TransE} can be selected as another baseline representing the methods of inferring a relation using only two entities.

\subsubsection{Experimental Results}

The evaluation results of path query answering for entity prediction are shown in Table \ref{table5}. We can discover that:
\begin{itemize}
  \item It is evident that our approach outperforms PTransE significantly on the three datasets. Specifically, our model achieves the performance gains: 26.3\%/ 30.9\%/ 74.6\% of MR/MRR/Hits@1 on FB15K, 55.0\%/ 5.3\%/ 10.3\% on WN18, and 6.5\%/4.2\%/5.1\% on NELL-995.
  \item Particularly, our model achieves satisfactory results on WN18; for instance, MR is 9 means any one of the top 10 candidate answers can be guaranteed as a correct answer. Hits@1 is 0.987 certifies that the correct rate of the only right answer is 98.7\%. These all illustrate the high accuracy of answering path queries for entity prediction with our model.
\end{itemize}

Table \ref{table6} reports the evaluation results of path query answering for relation inference. Some discussions are given:
\begin{itemize}
  \item The evaluation results demonstrate the superiority of employing multiple paths in path query answering for relation inference. TransE only calculates the score according to the translational distance between the embeddings of an entity pair in Eq.\ref{eq5a}, our model and PTransE introduce not only the translational distance between two entities but also the semantic similarity between the paths and the direct relation.
  \item Our model further outperforms PTransE consistently. The reason is that our model incorporates both entities and relations along paths to capture more semantics along paths; meanwhile, the horn rules and the entity converter are both to the benefit of representing paths with higher accuracy. This result illustrates that our approach can better represent and leverage paths in KG inference.
\end{itemize}

\subsubsection{Case Study}
To further illustrate the superiority of our model on inference with explainability, Table \ref{table7} provides three path query answering examples of relation inference on NELL-995 that are all capable of leveraging the horn rules to compose paths. On account of each query in our case study, an answer relation can be inferred by a horn rule matched to the path query. This answer relation even ranks first among all the candidate relations by the scores calculated by Eq.\ref{eq11}. More specific to Example \#2, given a path $jonathan \xrightarrow{has\_sibling} simon \xrightarrow{person\_born\_in\_city} york$ and a query $(jonathan, ?, york)$, a horn rule $person\_born\_in\_city (x, y) \Leftarrow has\_sibling (x, z) \wedge person\_born\_in\_city (z, y)$ can be matched to the path. The correct relation $person\_born\_in\_city$ for answering the query can be inferred by this rule to connect the head entity $jonathan$ and the tail entity $york$, which provides the human understandable explanations to achieving the answer. Furthermore, the confidence of this rule is 1 and it ensures the accuracy of the inference result. On the other hand, the rank of the correct relation is 1 among all the candidate relations calculated by Eq.\ref{eq11}, which also illustrates the good performance of our model on accurate path query answering.

\section{Conclusion}
\label{sec:conclusion}
This paper proposes a novel knowledge graph embedding model for inference via learning entity and relation embeddings by introducing a joint semantics and data-driven path representation. We compose the paths into single relations or condensed paths by the horn rules with high accuracy and explainability. For the paths that are still longer than 1, we develop a general entity converter strategy containing an attentive entity-to-type conversion module for the KG with types and an entity-to-relation space projection module for the KG without types to reduce the heterogeneity between entities and relations. Then, we design a path encoder to obtain the path representations with bi-directional path sequences input. We train our model to learn the entity and relation embeddings by introducing the fusion representation learning of multiple paths between each entity pair. Experimental results illustrate that our model achieves superior performance than the state-of-the-art baselines on link prediction and path query answering tasks. The detailed analyses also demonstrate the effectiveness of representing paths in a joint semantics data-driven fashion, making for the performance boost on KG inference tasks.


In terms of future work, we plan to further explore the following research directions: (1) One approach is to investigate the graph neural networks incorporated with horn rules to obtain the neighbor information to learn accurate KG embeddings. (2) A second direction is to provide further consideration to enhance the semantic representations of facts by including side information (e.g., textual descriptions of entities and ontology) beyond their relations that we have considered thus far.

\section{Compliance with Ethical Standards}

\pheadWithSpace{Funding} This work was partially funded by the National Natural Science Foundation of China (No.61772054), the NSFC Key Project (No.61632001), and a Fundamental Research Fund for the Central Universities of China.

\pheadWithSpace{Conflict of Interest} The authors declared that they have no conflicts of interest to the work in this article.

\pheadWithSpace{Ethical Approval} This article dose not contain any studies with human participants or animals performed by any of the authors.

\pheadWithSpace{Informed Consent} Informed consent was not required as no humans or animals were involved.

\printcredits

\bibliographystyle{cas-model2-names}

\bibliography{cas-refs}

\end{document}